\documentclass[letterpaper, 10 pt, conference]{ieeeconf}  

\IEEEoverridecommandlockouts                              

\usepackage{algorithm}
\usepackage{algpseudocode}
\usepackage{amsmath} 
\usepackage{amssymb}  
\usepackage{bm}
\usepackage{xspace}
\usepackage{xcolor}
\usepackage{graphicx} 
\usepackage{makecell}
\usepackage{booktabs}

\usepackage{etoolbox}

\makeatletter
\@ifundefined{ALG@step}{}{%
  \patchcmd{\ALG@step}{\stepcounter{ALG@line}}{\refstepcounter{ALG@line}}{}{}
  \providecommand*{\theHalgorithm}{\thealgorithm}
  \providecommand*{\theHALG@line}{}
  \renewcommand{\theHALG@line}{\theHalgorithm.\arabic{ALG@line}}
}
\makeatother

\usepackage{hyperref}
\usepackage{cleveref}

\newcommand{\solverName}[0]{Vectorized Online POMDP Planner\xspace}
\newcommand{\solverAbbr}[0]{VOPP\xspace}

\newcommand{\belSpace}{\ensuremath{\mathcal{B}}\xspace}
\newcommand{\bel}{\ensuremath{b}\xspace}

\newcommand{\belTree}{\ensuremath{\mathcal{T}}\xspace}

\newcommand{\stSpace}{\ensuremath{\mathcal{S}}\xspace}

\newcommand{\st}{\ensuremath{s}\xspace}
\newcommand{\stp}{\ensuremath{s'}\xspace}

\newcommand{\actSpace}{\ensuremath{\mathcal{A}}\xspace}
\newcommand{\act}{\ensuremath{a}\xspace}

\newcommand{\obsSpace}{\ensuremath{\mathcal{O}}\xspace}
\newcommand{\obs}{\ensuremath{o}\xspace}

\newcommand{\transF}{\ensuremath{T}\xspace}
\newcommand{\obsF}{\ensuremath{Z}\xspace}

\newcommand{\rewFunc}{\ensuremath{R}\xspace}

\newcommand{\pol}{\ensuremath{\pi}\xspace}
\newcommand{\optPol}{\ensuremath{\pi^*}\xspace}

\newcommand{\polTensor}{\ensuremath{\bm{\pi}}\xspace}

\DeclareMathOperator{\reals}{\mathbb{R}}

\DeclareMathOperator*{\argmax}{arg\,max}
\newcommand{\expect}{\mathbb{E}}

\newcommand{\tensor}[1]{\ensuremath{\mathbf{#1}}\xspace}

\newcommand{\numParallel}{\ensuremath{n_p}\xspace}

\newcommand{\keywordcolor}{black}
\algrenewcommand\algorithmicfor{\textcolor{\keywordcolor}{\textbf{for}}}
\algrenewcommand\algorithmicdo{\textcolor{\keywordcolor}{\textbf{do}}}
\algrenewcommand\algorithmicend{\textcolor{\keywordcolor}{\textbf{end}}}

\title{\LARGE \bf
Vectorized Online POMDP Planning
}

\author{Marcus Hoerger, Muhammad Sudrajat, Hanna Kurniawati
\thanks{*This work was supported by the ARC Research Hub in Intelligent Robotic Systems for Real-Time Asset Management (IH210100030), in collaboration with the University of Sydney and Nexxis Technology.}%
\thanks{The authors are with the School of Computing, Australian National University, Australia 
{\tt\small \{marcus.hoerger, muhammad.sudrajat, hanna.kurniawati\}@anu.edu.au}}%
}

\begin{document}

\maketitle
\thispagestyle{empty}
\pagestyle{empty}

\begin{abstract}
Planning under partial observability is an essential capability of autonomous robots. The Partially Observable Markov Decision Process (POMDP) provides a powerful framework for planning under partial observability problems, capturing the stochastic effects of actions and the limited information available through noisy observations. POMDP solving could benefit tremendously from massive parallelization on today's hardware, but parallelizing POMDP solvers has been challenging. Most solvers rely on interleaving numerical optimization over actions with the estimation of their values, which creates dependencies and synchronization bottlenecks between parallel processes that can offset the benefits of parallelization. In this paper, we propose \solverName (\solverAbbr), a novel parallel online solver that leverages a recent POMDP formulation which analytically solves part of the optimization component, leaving numerical computations to consist of only estimation of expectations. \solverAbbr represents all data structures related to planning as a collection of tensors, and implements all planning steps as fully vectorized computations over this representation. The result is a massively parallel online solver with no dependencies or synchronization bottlenecks between concurrent processes. Experimental results indicate that \solverAbbr is at least $20\times$ more efficient in computing near-optimal solutions compared to an existing state-of-the-art parallel online solver. Moreover, \solverAbbr outperforms state-of-the-art sequential online solvers, while using a planning budget that is $1000\times$ smaller.   



\end{abstract}


\section{INTRODUCTION}
Planning under partial observability is an essential, yet challenging problem for autonomous robots. The Partially Observable Markov Decision Process (POMDP)~\cite{kaelbling1998planning} is a principled framework to solve planning under uncertainty problems. It lifts the planning problem from the robot's state space to its belief space, the space of all probability distributions over the state space. Although solving POMDPs exactly is computationally intractable in general~\cite{papadimitriou1987complexity}, many scalable approximately optimal online solvers have been proposed (reviewed in~\cite{Kur22:Partially}), and some have been applied to realistic robot applications, such as~\cite{hoerger2019candy,bai2012unmanned,saleem2024pomdp,lauri2022partially}. 

However, most POMDP solvers do not exploit the massive parallelisation that Graphics Processor Units (GPU) offer. Paralellising POMDP solving is quite involved. POMDP solving requires interleaving of numerical optimisation to find actions with the highest expected total reward and estimation of expected total rewards themselves. When parallelised, this interleaving process creates dependencies that make load balancing difficult. 
Careful parallelisation strategies have been proposed for POMDP solving, including process synchronization and scheduling~\cite{wray2015parallel,lee2016massively,basu2021parallelizing,Pai21HyPdespot}. They have significantly improved the scalability of serial approximate POMDP solvers, but incur substantial process coordination overhead that limits the benefits of massive parallelisation. In contrast, our method builds on a recent approach to approximate POMDPs solutions~\cite{kim2025porpp} that partially solve the optimisation component analytically, leaving numerical computation only for estimation of expectations.


Specifically, we propose \solverName (\solverAbbr), a new parallel online POMDP solver based on PORPP~\cite{kim2025porpp}.
Similar to most online solvers, \solverAbbr is a tree search-based method. Starting from the current belief, we perform guided belief space sampling followed by backup operations to construct a representative belief tree and evaluate different action sequences. However, unlike existing solvers, \solverAbbr represents all data structures associated with the belief tree as a collection of tensors. Crucially, both guided belief space sampling and backup are implemented as \textit{fully vectorized} computations over this tensor representation. This enables \solverAbbr to fully harness the immense data-parallel throughput of modern GPUs. The result is a massively parallel online POMDP solver---running entirely on the GPU---that uses tens of thousands of parallel simulations to compute a policy, with no explicit synchronization between simulations required.



To the best of our knowledge, \solverAbbr is the first fully vectorized online POMDP solver. 
Experimental results on three POMDP benchmark problems indicate that \solverAbbr is at least $20\times$ more efficient in computing near-optimal policies compared to the current state-of-the-art parallel online solver, HyP-DESPOT~\cite{Pai21HyPdespot}, for problems with large state, action, and observation spaces; for some benchmarks, \solverAbbr is more than $100\times$ faster than HyP-DESPOT. \solverAbbr is open source and available at \url{https://github.com/RDLLab/VOPP}.



\section{BACKGROUND AND RELATED WORK}

\subsection{Partially Observable Markov Decision Process (POMDP)}\label{ssec:background_pomdp}
A POMDP provides a general mathematical framework for sequential decision-making under uncertainty.
Formally, a POMDP is an 8-tuple $\mathcal{P} = \left\langle \stSpace, \actSpace, \obsSpace, \transF, \obsF, \rewFunc, \bel_{0}, \gamma \right\rangle$. Initially, the robot is in a hidden state $s_{0} \in \stSpace$. This uncertainty is represented by an initial belief $\bel_0 \in \belSpace$, a probability distribution on the state space $\stSpace$, where \belSpace is the set of all possible beliefs. At each step $t \ge 0$, the robot executes an action $\act_{t} \in \actSpace$ according to some policy $\pol$. Due to stochastic effects of executing actions, it transitions from the current state $\st_t\in\stSpace$ to a next state $\st_{t+1} \in \stSpace$ according to the transition model $\transF(\st_{t}, \act_{t}, \st_{t+1}) = p(\st_{t+1} \mid \st_{t}, \act_{t})$, which is a conditional probability function. The robot does not know the state $\st_{t+1}$ exactly, but perceives an observation $\obs_{t} \in\obsSpace$ from the environment according to the observation model $\obsF(\st_{t+1}, \act_{t}, \obs_{t}) = p(\obs_{t} \mid \st_{t+1}, \act_{t})$. 
In addition, the robot receives an immediate reward $r_{t} = R(\st_{t}, \act_{t}) \in \reals$. The goal is to find a policy $\pol$ that maximizes the expected total discounted reward or the {\em policy value}
\begin{align}
    \mathcal{V}_{\pol}(\bel_{0}) = \expect\left[\sum_{t=0}^{\infty}\gamma^t r_t \, \bigg\vert\, \bel_{0}, \pi\right],
\end{align}
where the discount factor $0 < \gamma < 1$ ensures that $\mathcal{V}_{\pol}(\bel)$ is finite and well-defined.

The robot's decision space is the set $\Pi$ of policies, defined as mappings from beliefs to actions. In this paper, we consider {\em stochastic policies}, i.e., $\pol$ is a belief-dependent distribution over the action space. The POMDP solution is then the optimal policy, denoted as \optPol and given by 
\begin{align}
\optPol = \argmax_{\pol \in \Pi} \mathcal{V}_{\pol}(\bel),
\end{align}
for each belief $\bel\in\belSpace$.
A more elaborate explanation is available in~\cite{kaelbling1998planning}.

\subsection{Parallel Planning under Uncertainty}\label{background:parallel_pomdp}
Advances in multicore CPUs and GPUs have enabled parallelization of online planning under uncertainty methods. Most approaches focus on solving MDPs -- the fully observable variant of POMDPs -- and build on Monte Carlo Tree Search (MCTS). The works in~\cite{cazenave2007parallelization, chaslot2008parallel} focus on different MCTS parallelization approaches, including leaf, root, and tree parallelization. Leaf parallelization evaluates leaf nodes using parallel simulations; root parallelization constructs multiple trees in parallel and merges them after search; and tree parallelization runs concurrent searches within a single tree, requiring extensive mutex-based synchronization. 

For POMDPs, several parallel offline solvers have been proposed. The work in~\cite{wray2015parallel} proposed an offline solver based on Point-Based Value Iteration (PBVI)~\cite{Pin03:Pointt} implemented on GPUs to accelerate the backup step by exploiting sparsity in belief vectors and optimizing memory access. The work in~\cite{lee2016massively} introduces offline solvers based on Monte Carlo Value Iteration (MCVI)~\cite{bai2010monte} that exploit GPU-only and hybrid CPU–GPU architectures to parallelize action evaluation, belief node value estimation, and expected return computation.

More recently, parallel online solvers have been developed, which aim to parallelize existing tree search-based methods. The work in~\cite{basu2021parallelizing} proposes a parallelized version of POMCP~\cite{silver2010montee}, using root parallelization and pursuing tree parallelization, to speed up the action selection in large POMDPs. The online solver HyP-DESPOT~\cite{Pai21HyPdespot} proposes a CPU–GPU hybrid architecture to parallelize the belief tree search on the CPU and Monte Carlo simulations on the GPU, combining them in a hybrid architecture. 

Although these online solvers demonstrate remarkable speed-ups compared to their serial counterparts, they require careful synchronization between parallel simulations to ensure consistent updates of belief-tree statistics such as visitation counts and value estimates. This limits their scalability, as excessive synchronization overhead can quickly offset the benefits of parallelism. Moreover, to ensure that parallel simulations explore the belief tree sufficiently, they rely on auxiliary mechanisms such as virtual losses~\cite{Pai21HyPdespot}, which complicates their implementation and biases the search.

In contrast, \solverAbbr is a fully vectorized online solver running entirely on the GPU. It requires neither synchronization between parallel computations nor any CPU--GPU data exchange. This significantly simplifies \solverAbbr's architecture and allows us to fully exploit the massive data parallel throughput of modern GPUs.


\section{\solverName}
In this section, we present our fully vectorized solver \solverName (\solverAbbr). In this context, vectorization refers to the reformulation of all computational steps as batched operations on tensors, a practice that aligns with the Single Instruction, Multiple Data (SIMD) paradigm of GPUs. For completeness, we first provide a brief overview of PORPP in \Cref{ssec:porpp}, followed by the description of \solverAbbr in \Cref{ssec:solver,ssec:forward_search,ssec:tensor_data_structure,ssec:backup}.

\subsection{PORPP}\label{ssec:porpp}
Partially Observable Reference Policy Programming (PORPP)~\cite{kim2025porpp} is a recently proposed online POMDP solver. It builds on the concept of {\em Reference-Based} POMDPs~\cite{kim2023reference}, a reformulation of a POMDP whose {\em analytical} objective is the POMDP value function, penalized by the Kullback-Leibler (KL) divergence between the maximizing policy and a user-defined \textit{reference policy} $\pi_0$. For a belief $\bel\in\belSpace$, this objective can be compactly written as
\begin{equation}\label{eq:analytical_belief_value}
\mathcal{V}(\bel) = \frac{1}{\eta}\log\Big[ \sum_{\act\in\actSpace}\exp[\eta\Psi(\bel, \act)]\Big] := [\mathcal{L}_{\eta}\Psi](\bel),
\end{equation}
where $\eta > 0$ is a temperature parameter, balancing reward maximization with deviation from the reference policy, and $\mathcal{L}_{\eta}$ is the log-sum-exp operator~\cite{blanchard2021accurately}. The expression $\Psi$ denotes \textit{preferences} over belief-action pairs:  
\begin{align}\label{eq:preferences}
\Psi(\bel, \act) &= \frac{1}{\eta}\log(\pol_0(\act \mid \bel)) + \mathcal{R}(\bel, \act) \\ \nonumber
&\qquad\qquad + \gamma \sum_{\obs\in\obsSpace} p(\obs \mid \bel, \act)[\mathcal{L}_{\eta}\Psi](\tau(\bel,\act,\obs)),
\end{align}
where $\mathcal{R}(\bel, \act)$ is a Monte Carlo estimate of $\int_{\st\in\stSpace}R(s, a)\bel(s)\mathrm{d}s$ and $\tau$ is the belief update operator.

For many POMDP problems, it is possible to design a reference policy $\pol_0$ that encodes domain knowledge. For instance, for motion planning under uncertainty problems, the approach in~\cite{liang2024scaling} samples (macro)-actions from the reference policy using a fast deterministic sampling-based motion planner~\cite{thomason2024motions}. If such domain knowledge is unavailable, $\pol_0$ can be a uniform distribution over the action space, corresponding to uniformly initialized action preferences. In our experiments in \Cref{sec:experiments}, we use uniform initial reference policies.

To correct any misspecifications of $\pol_0$, PORPP proposes an iterative scheme which gradually deforms $\pol_0$ towards an optimal policy $\pi^*$ for the original POMDP. This is done by iteratively updating the preferences in \cref{eq:preferences} via
\begin{align}\label{eq:preference_update}
\Psi_{k+1}(\bel, \act) &= \Psi_k(\bel, \act) - [\mathcal{L}_{\eta}\Psi_k](\bel)\\ \nonumber
&+ \mathcal{R}(\bel, \act) + \gamma\sum_{\obs\in\obsSpace}p(\obs \mid \bel, \act)[\mathcal{L}_{\eta}\Psi_{k}](\tau(\bel, \act, \obs)),
\end{align}
where the policy at iteration $k$ is derived from the preference values via the softmax function
\begin{equation}\label{eq:ref_policy_k}
\pol_k(\bel, \act) = \frac{\exp[\eta\Psi_k(\bel,\act)]}{\sum_{\act'\in\actSpace}\exp[\eta\Psi_k(\bel, \act')]}.
\end{equation}
In practice, PORPP interleaves belief space sampling with preference backups to approximate the action preference updates in \cref{eq:preference_update}. The sampled beliefs are maintained in a belief tree $\mathcal{T}$, consisting of belief and action nodes. Each belief node branches into action nodes, while each action node branches into successor belief nodes based on sampled observations. PORPP constructs \belTree by iterating the following steps:
\begin{enumerate}
    \item Forward search: Starting from the current belief $\bel_0$, PORPP samples an \textit{episode}, i.e., a sequence of state–action–observation–reward quadruples up to a maximum depth, and adds new belief nodes along the episode's action–observation history if they do not exist yet. At each visited belief $\bel\in\belSpace$, PORPP samples an action from the current reference policy, \cref{eq:ref_policy_k}, associated with the belief. 
    \item Preference backup: After sampling an episode, PORPP traverses the sequence of visited beliefs back to the root and updates the preferences of the sampled actions according to \cref{eq:preference_update} at each visited belief $\bel$.
\end{enumerate}
PORPP repeats these steps until the planning budget for the current step is exceeded. 

Many online solvers select actions according to some variant of UCT~\cite{kocsis2006bandit} during the forward search, which requires maximization over action values at each belief node. In contrast, PORPP selects actions by sampling from the current reference policy, which is embarrassingly parallel. Moreover, PORPP computes belief values analytically according to \cref{eq:analytical_belief_value}. Both operations---embarrassingly parallel action {\em sampling} and {\em analytical} belief value computation---open up new avenues for efficient parallelization, which we exploit with \solverAbbr.

\subsection{\solverAbbr Overview}\label{ssec:solver}
\begin{figure*}[htbp]
  \centering
  \begin{tabular}{cc}
    \includegraphics[height=0.25\textwidth]{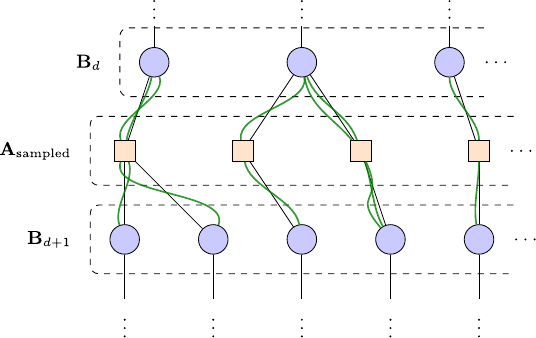} &
    \includegraphics[height=0.25\textwidth]{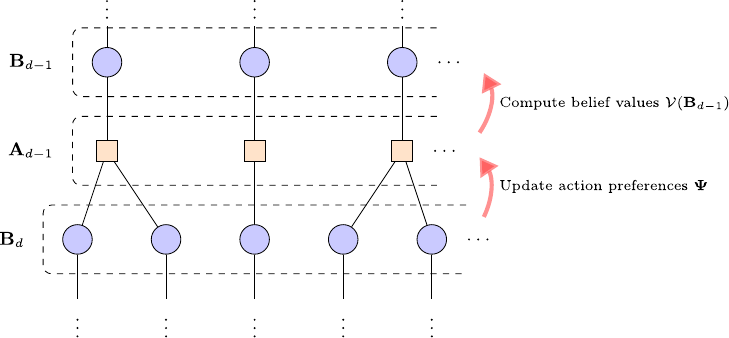} \\
    (a) Vectorized forward search & (b) Vectorized preference backup
  \end{tabular}
  \caption{Illustration of the two vectorized main operations---forward search (a) and preference backup (b)---of \solverAbbr. Blue circles represent belief nodes, while yellow squares represent action nodes. The green lines represent sampled episodes. (a) {\em Vectorized forward search}: \solverAbbr samples an action for each episode from the belief nodes $\tensor{B}_{d}$ at depth $d$ in parallel and collects the sampled actions in the action tensor $\tensor{A}_{\text{sampled}}$. It then performs a vectorized forward simulation of the episodes from one step using the generative model $G$ and $\tensor{A}_{\text{sampled}}$. For the resulting observations, \solverAbbr appends new belief nodes to \tensor{B} if they do not exist yet. The search then continues from the belief nodes at depth $d+1$ that the episodes visit. (b) {\em Vectorized preference backup}: For all belief nodes $\tensor{B}_d$ at depth $d$, \solverAbbr updates the preference values $\tensor{\Psi}$ of their parent actions in single vectorized step. The updated preference values are then used to compute the belief values of all beliefs $\tensor{B}_{d-1}$ at depth $d-1$ in one vectorized step, before the backup continues from $d-1$.}
  \label{fig:vec_operations}
\vspace{-10pt}
\end{figure*}


\begin{algorithm}
\caption{\textsc{\solverAbbr}}
\label{alg:solver}
\begin{algorithmic}[1]
\Require Initial belief $\bel$, Num. parallel simulations $\numParallel$, Temperature $\eta$
\While{Problem not terminated}
\State $\belTree \gets$ Initialize tensors $\tensor{B}, \tensor{A}, \tensor{\Psi}$
\State $D_{\max} \gets 1$
\While{planning budget not exceeded}\label{line:planning_loop_start}
    \State $d \gets 0$
    \State $\tensor{S} \gets \textsc{SampleStatesFromBelief}(\bel, \numParallel)$ \label{line:sample_states}
    \State $\tensor{B}_{\text{curr}} \gets$ Root node indices of size $\left |\tensor{S} \right |$ \label{line:make_curr_bel}
    \State $(\tensor{B}_{\text{leaf}}, \tensor{H}) \gets \textsc{Search}(\belTree, \tensor{B}_{\text{curr}}, \tensor{S}, d, D_{\max})$ \label{line:simulate} \Comment{Alg.~\ref{alg:simulate}}
    \State $\belTree \gets \textsc{Backup}(\belTree, \tensor{B}_{\text{leaf}}, \tensor{H}, D_{\max})$ \label{line:backup} \Comment{Alg.~\ref{alg:backup}}
    \State $D_{\max} \gets D_{\max} + 1$
\EndWhile \label{line:planning_loop_end}
\State $\tensor{B}_0 \gets$ Root node in \belTree
\State $\act \gets \argmax_{\act} \tensor{\Psi}(\tensor{B}_0, \act)$ \label{line:act_select}
\State Execute action $\act$ and perceive observation $o$\label{line:exec_action}
\State $\bel \gets \tau(\bel, \act, \obs)$\label{line:update_belief}
\EndWhile
\end{algorithmic}
\end{algorithm}
\solverAbbr is an anytime parallel online POMDP solver based on PORPP. Key to \solverAbbr is representing all data structures associated with the belief tree $\belTree$ as tensors. This allows \solverAbbr to implement PORPP's key steps---{\em forward search} and {\em preference backup}---as fully vectorized computations manipulating the tensor data structures of \belTree. In contrast to existing parallel online POMDP solvers, this design requires no synchronization between concurrent computations, enabling \solverAbbr to achieve a significantly higher computational throughput on massively parallel hardware, such as GPUs.

Suppose the POMDP to be solved is $\mathcal{P} = \left\langle \stSpace, \actSpace, \obsSpace, \transF, \obsF, \rewFunc, \bel_{0}, \gamma \right\rangle$.
The state space \stSpace, action space \actSpace, and observation space \obsSpace can be discrete, continuous, or hybrid. For continuous action/observation spaces, we represent the space with a fixed, yet representative set of sampled actions/observations with finite size, which is selected a priori. We assume access to a stochastic generative model $G: \stSpace\times\actSpace \mapsto \stSpace\times\obsSpace\times\mathbb{R}$ to simulate the transition, observation, and reward models. For a state $\st\in\stSpace$ and action $\act\in\actSpace$, the model $G$ produces a next state $\stp\in\stSpace$, observation $\obs\in\obsSpace$ and reward $r \in \mathbb{R}$, such that $(\stp, \obs)$ is distributed according to $p(\obs, \stp \mid \st, \act) = \transF(\st, \act, \stp)\obsF(\stp, \act, \obs)$, and $r=\rewFunc(\st, \act)$. We further assume that $G$ is implemented as a vectorized model, i.e., for a tensor of states and actions, it produces a tensor of next states, observations, and rewards. In the remainder of the paper, we use \textbf{BOLD} upper-case letters to denote tensors. 

The key steps of \solverAbbr are presented in \Cref{alg:solver}. At each planning loop iteration, (\crefrange{line:planning_loop_start}{line:planning_loop_end}), \solverAbbr performs a vectorized forward search to expand the belief tree by one level, followed by a vectorized backup operation (\cref{line:backup}) to update the action preferences stored in \belTree. Details on the vectorized forward search and backup operations are provided in \Cref{ssec:forward_search} and \Cref{ssec:backup}, respectively. \Cref{fig:vec_operations} provides an illustration of both steps. They repeat until the planning budget for the current step is exceeded. Finally, we select the action with the highest preference value at the root node (\cref{line:act_select}), execute the action in the environment (\cref{line:exec_action}) and update the belief based on the action executed and observation perceived (\cref{line:update_belief}). To update the belief, we use a Sequential Importance Resampling particle filter~\cite{arulampalam2002tutorial}. The next section details our belief tree tensor data structure.



\subsection{Belief Tree Tensor Data Structure}\label{ssec:tensor_data_structure}
To enable a fully vectorized implementation of \solverAbbr, we represent all internal data structures associated with the belief tree \belTree as a collection of three tensors, \tensor{B}, \tensor{A}, and \tensor{\Psi}. The tensors $\tensor{B}$ and $\tensor{A}$ represent the belief and action nodes, including their parent-child relations. Specifically, \tensor{B} is a 2D tensor, where each row corresponds to a belief node and contains two entries: the first stores the row index of the parent action node in \tensor{A}, while the second stores the parent observation. For the root node, stored in the first row of $\tensor{B}$, both entries are set to \texttt{NULL}. The tensor $\tensor{A}$ is a 2D tensor in which each row represents an action node and contains four entries: The first entry stores the row index in $\tensor{B}$ of the node’s parent belief, while the second stores the action associated with the node. The final two entries store the action node's cumulative immediate reward and visitation count, respectively, which are updated during the forward search. Finally, $\tensor{\Psi}$ is a 2D tensor with $1 + |\actSpace|$ columns. Each row stores the current action preference values (defined in \cref{eq:preferences}) for a specific belief. The first column stores the row index in $\tensor{B}$ corresponding to that belief, while the remaining $|\actSpace|$ columns store the preference value for each action. 

Together, these tensors form a compact representation of the belief tree, $\belTree = \{\tensor{B}, \tensor{A}, \tensor{\Psi}\}$, which enables fully vectorized forward search and backup operations, as detailed in the next two sections.

\subsection{Vectorized Forward Search}\label{ssec:forward_search}
\begin{algorithm}[!htp]
\caption{\textsc{Search}$(\belTree, \tensor{B}_{\text{curr}}, \tensor{S}, d, D_{\max})$}
\label{alg:simulate}
\begin{algorithmic}[1]
\If{$d > D_{\max}$}
  \State \Return $(\tensor{B}_{\text{curr}}, \textsc{ValueHeuristic}(\tensor{S}))$\label{line:value_heuristic}
\EndIf
\State $\polTensor(\tensor{B}_{\text{curr}}, \cdot) \gets \textsc{Softmax}[\eta \cdot \tensor{\Psi}(\tensor{B}_{\text{curr}}, \cdot)]$ \label{line:softmax_pol}
\State $\tensor{A}_{\text{sampled}} \sim \polTensor(\tensor{B}_{\text{curr}}, \cdot)$ \label{line:make_action_tensor}

\State $(\tensor{S}', \tensor{O}, \tensor{R}) \gets G(\tensor{S}, \tensor{A}_{\text{sampled}})$ \label{line:forward_step}\Comment{Generative model}
\State $(\tensor{A}_{\text{node}}, \belTree) \gets \textsc{AppendActions}(\belTree, \tensor{B}_{\text{curr}}, \tensor{A}_{\text{sampled}}, \tensor{R})$ \label{line:append_actions}
\State $(\tensor{A}_{node}, \tensor{O}, \tensor{S}') \gets \textsc{FilterTerminals}(\tensor{A}_{node}, \tensor{O}, \tensor{S}')$ \label{line:filter_terminals}
\State $(\tensor{B}_{\text{next}}, \belTree) \gets \textsc{AppendBeliefs}(\belTree, \tensor{A}_{\text{node}}, \tensor{O})$\label{line:append_beliefs}
\State \Return $\textsc{Search}(\belTree, \tensor{B}_{\text{next}}, \tensor{S}', d + 1, D_{\max})$\label{line:next_simulation}
\end{algorithmic}
\end{algorithm}

\Cref{alg:simulate} presents the pseudocode for \solverAbbr's vectorized forward search. At each planning loop iteration, we first sample a batch of size \numParallel of initial states from the current belief and store them in a state tensor \tensor{S} (\cref{line:sample_states} in \Cref{alg:solver}). The parameter \numParallel determines the number of episodes that are sampled in parallel (in our experiments, we use up to $60,000$ parallel episodes). We also construct a matching index tensor $\tensor{B}_{\text{curr}}$ of the same batch size (\cref{line:make_curr_bel}), where each entry points to the root node in $\tensor{B}$, thereby associating each sampled state in $\tensor{S}$ with the initial belief node in the tree. We then recursively sample a batch of episodes, i.e., sequences of state–action–observation–reward quadruples, starting from the initial states in \tensor{S}, as follows: 

For each belief indexed by $\tensor{B}_{\text{curr}}$, we first construct a softmax policy $\pol$ (\cref{line:softmax_pol}) using the corresponding action preferences in $\tensor{\Psi}(\tensor{B}_{\text{curr}}, \cdot)$ according to \cref{eq:ref_policy_k}. We then sample an action for each belief from this newly constructed policy and store the results in the action tensor $\tensor{A}_{\text{sampled}}$ (\cref{line:make_action_tensor}). Note that both the policy construction and action sampling steps are fully vectorized over all entries in $\tensor{B}_{\text{curr}}$. The batch of states in $\tensor{S}$ and corresponding actions in $\tensor{A}_{\text{sampled}}$ are then simulated forward in a single vectorized step using the generative model $G$, yielding the next state tensor $\tensor{S}'$, the observation tensor $\tensor{O}$, and the reward tensor $\tensor{R}$ (\cref{line:forward_step}).

Following the forward simulation step, the belief tree is expanded with the sampled actions and observations (\crefrange{line:append_actions}{line:append_beliefs}). To ensure that no action or belief node is added more than once, we use the following vectorized expansion process: First, we pair each belief index in $\tensor{B}_{\text{curr}}$ with the corresponding action in $\tensor{A}_{\text{sampled}}$ to form a batch of belief-action pairs and identify all unique pairs. Using a fast hash-based matching algorithm, we efficiently determine which action nodes corresponding to these unique pairs already exist in \tensor{A}. New, previously unseen pairs are concatenated to $\tensor{A}$, and a single index tensor $\tensor{A}_{\text{node}}$ is returned. This tensor provides the corresponding action node index in \tensor{A} (either existing or newly created) for each entry in the sampled action tensor $\tensor{A}_{\text{sampled}}$. During this process, the cumulative rewards and visit counts stored in $\tensor{A}$ are updated for all affected action nodes in a single vectorized step. To ensure that terminated episodes do not further contribute to the forward search, we filter rows in $\tensor{A}_{\text{node}}$, $\tensor{O}$ and $\tensor{S}'$ that correspond to terminated episodes (\cref{line:filter_terminals}).

To append new belief nodes to \tensor{B}, we use a similar vectorized procedure. We pair each action node index in $\tensor{A}_{\text{node}}$ with the corresponding observation in $\tensor{O}$ and identify unique action-observation pairs. These pairs are then matched against existing belief nodes in $\tensor{B}$, with new nodes created as needed. This returns an index tensor $\tensor{B}_{\text{next}}$, pointing to the belief nodes for the next simulation step (\cref{line:next_simulation}).

This recursive forward search continues until depth $D_{\max}$. For the resulting belief nodes at depth $D_{\max}$, a state-based, problem-dependent heuristic function estimates their values from $\tensor{S}$ (\cref{line:value_heuristic}). These estimates serve as the initial values for the subsequent backup phase.

\subsection{Vectorized Preference Backup}\label{ssec:backup}
\vspace{-5pt}
\begin{algorithm}
\caption{\textsc{Backup}$(\belTree, \tensor{B}_{\text{leaf}}, \tensor{H}, D_{\max})$}
\label{alg:backup}
\begin{algorithmic}[1]
\State $\tensor{N}(\tensor{B}_{\text{leaf}}) \gets \textsc{AggregateBeliefVisits}(\tensor{B}_{\text{leaf}})$\label{line:algo_3_aggregate_visits}
\State $\tensor{C}(\tensor{B}_{\text{leaf}}) \gets \textsc{AggregateHeuristics}(\tensor{B}_{\text{leaf}}, \tensor{H})$\label{line:aggregate_heuristics}
\State $\mathcal{V}(\tensor{B}_{\text{leaf}}) \gets \tensor{C}(\tensor{B}_{\text{leaf}}) / \tensor{N}(\tensor{B}_{\text{leaf}})$\label{line:calc_leaf_values}

\For{$d = D_{\max}, D_{\max} - 1, \dotsc, 1$}  
  \State Let $\tensor{B}_d$ be the tensor of belief nodes at depth $d$ in $\belTree$
  \State Let $\tensor{A}_{d-1}$ be the tensor of parent actions of $\tensor{B}_d$ in $\belTree$
  \State Let $\tensor{B}_{d-1}$ be the tensor of parent beliefs of $\tensor{A}_{d-1}$ in \belTree
  \State $\tensor{R}(\tensor{B}_{d-1}, \tensor{A}_{d-1})\gets \frac{\tensor{C}(\tensor{B}_{d-1}, \tensor{A}_{d-1})}{\tensor{N}(\tensor{B}_{d-1}, \tensor{A}_{d-1})}$\label{line:R_tensor}
  \State $\tensor{W}(\tensor{A}_{d-1}) \gets \textsc{WeightedSum}(\mathcal{V}(\tensor{B}_d), N(\tensor{B}_d))$\label{line:W_tensor}
  \State $\tensor{Q}(\tensor{B}_{d-1}, \tensor{A}_{d-1}) \gets \tensor{R}(\tensor{B}_{d-1}, \tensor{A}_{d-1}) + \gamma \tensor{W}(\tensor{A}_{d-1})$\label{line:Q_tensor}
  \State $\tensor{N}(\tensor{B}_{d-1}) \gets \textsc{SumActionVisits}(\tensor{N}(\tensor{B}_{d-1}, \tensor{A}_{d-1}))$\label{line:visit_count_aggr}
  \State $\mathcal{V}_{\text{curr}}(\tensor{B}_{d-1}) \gets [\mathcal{L}_{\eta}\tensor{\Psi}](\tensor{B}_{d-1})$\label{line:v_curr}
  \State $\tensor{\Psi}(\tensor{B}_{d-1}, \cdot) \gets \tensor{\Psi}(\tensor{B}_{d-1}, \cdot) - \mathcal{V}_{\text{curr}}(\tensor{B}_{d-1}) + \tensor{Q}(\tensor{B}_{d-1}, \tensor{A}_{d-1})$\label{line:psi_update_tensor}
\State $\mathcal{V}(\tensor{B}_{d-1}) \gets [\mathcal{L}_{\eta}\tensor{\Psi}](\tensor{B}_{d-1})$\label{line:v_final}
\EndFor
\end{algorithmic}
\end{algorithm}

After sampling a batch of episodes as described in the previous section, we perform a sequence of vectorized backup operations to update the action preferences values at the sampled beliefs (\Cref{alg:backup}). 

The backup process begins at the leaf nodes reached by the sampled episodes. We first perform vectorized aggregation operations to initialize their values based on the heuristic estimates in $\tensor{H}$. The function $\textsc{AggregateBeliefVisits}$ (\cref{line:algo_3_aggregate_visits} in \Cref{alg:backup}) performs a batched count of each unique belief node index in the leaf node tensor $\tensor{B}_{\text{leaf}}$ to determine their visit counts. Similarly, $\textsc{AggregateHeuristics}$ (\cref{line:aggregate_heuristics}) performs a batched sum of the heuristic values $\tensor{H}$ for each of these unique leaf nodes. The leaf value estimates $\mathcal{V}(\bel)$ are then computed from $\tensor{H}$ and the visit counts (\cref{line:calc_leaf_values}).

The backup then proceeds iteratively from the leaf nodes $d=D_{\max}$ to the root. We perform a series of vectorized computations on {\em all} nodes at a given depth to update the corresponding action preferences \tensor{\Psi} according to \cref{eq:preference_update}.

First, we compute a tensor of $Q$ values for all action nodes $\tensor{A}_{d-1}$ (\crefrange{line:R_tensor}{line:Q_tensor}). These $Q$ values combine the average immediate reward $\tensor{R}$, computed from the cumulative rewards $\tensor{C}(\tensor{B}_{d-1}, \cdot)$ and the visit counts $\tensor{N}(\tensor{B}_{d-1}, \cdot)$ stored in the global action tensor $\tensor{A}$ (\cref{line:R_tensor}), with the expected future value $\tensor{W}$. 
This future value is computed by the $\textsc{WeightedSum}$ function (\cref{line:W_tensor}) as a visit-count–weighted average of the values of all child belief nodes in $\tensor{B}_d$. Importantly, $\textsc{WeightedSum}$ uses the action’s total visit count as the normalizer; thus, episodes that terminate after selecting the action (and therefore do not create a child belief node) do not contribute to the future-value term.

With the $Q$ values computed for all actions at depth $d-1$, we update their action preference values and the values of their parent belief nodes in $\tensor{B}_{d-1}$. The visit counts for these parent nodes are computed by aggregating their corresponding child action visits (\cref{line:visit_count_aggr}). The action preferences $\tensor{\Psi}$ are updated using the newly computed $Q$ values. To do this, we compute the current values $\mathcal{V}_{\text{curr}}$ of the beliefs in $\tensor{B}_{d-1}$ using the existing preference values and the log-sum-exp operator (\cref{line:v_curr}). With the current belief values and the computed $Q$ values, we update the action preferences $\tensor{\Psi}$ (\cref{line:psi_update_tensor}). Finally, the value $\mathcal{V}(\bel)$ for each belief node in $\tensor{B}_{d-1}$ is recalculated from these updated preferences (\cref{line:v_final}), before the backup progresses with the next iteration.

\section{EXPERIMENTS AND RESULTS}\label{sec:experiments}
\begin{figure}[htbp]
  \centering
  \begin{tabular}{@{}c@{}c@{}c@{}}
    \includegraphics[height=0.159\textwidth]{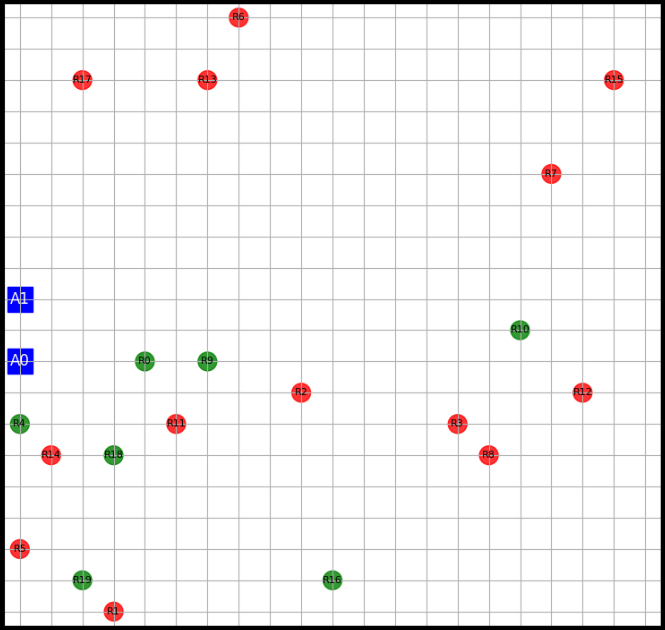}  &
    \includegraphics[height=0.159\textwidth]{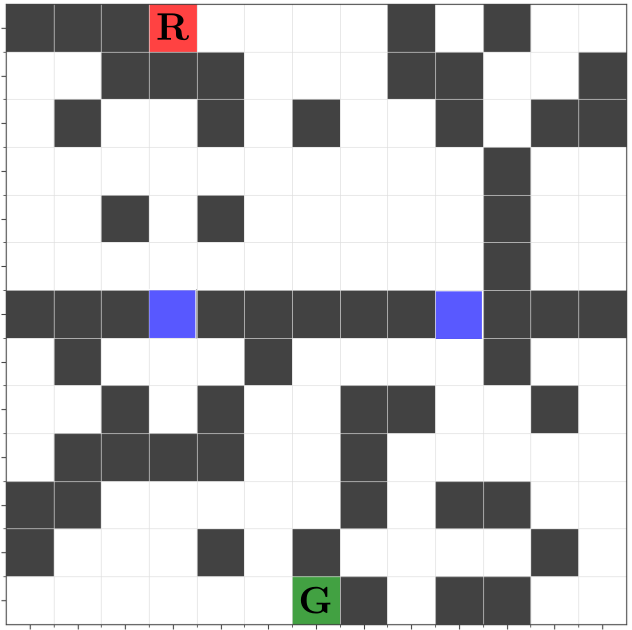} &
    \includegraphics[height=0.159\textwidth]{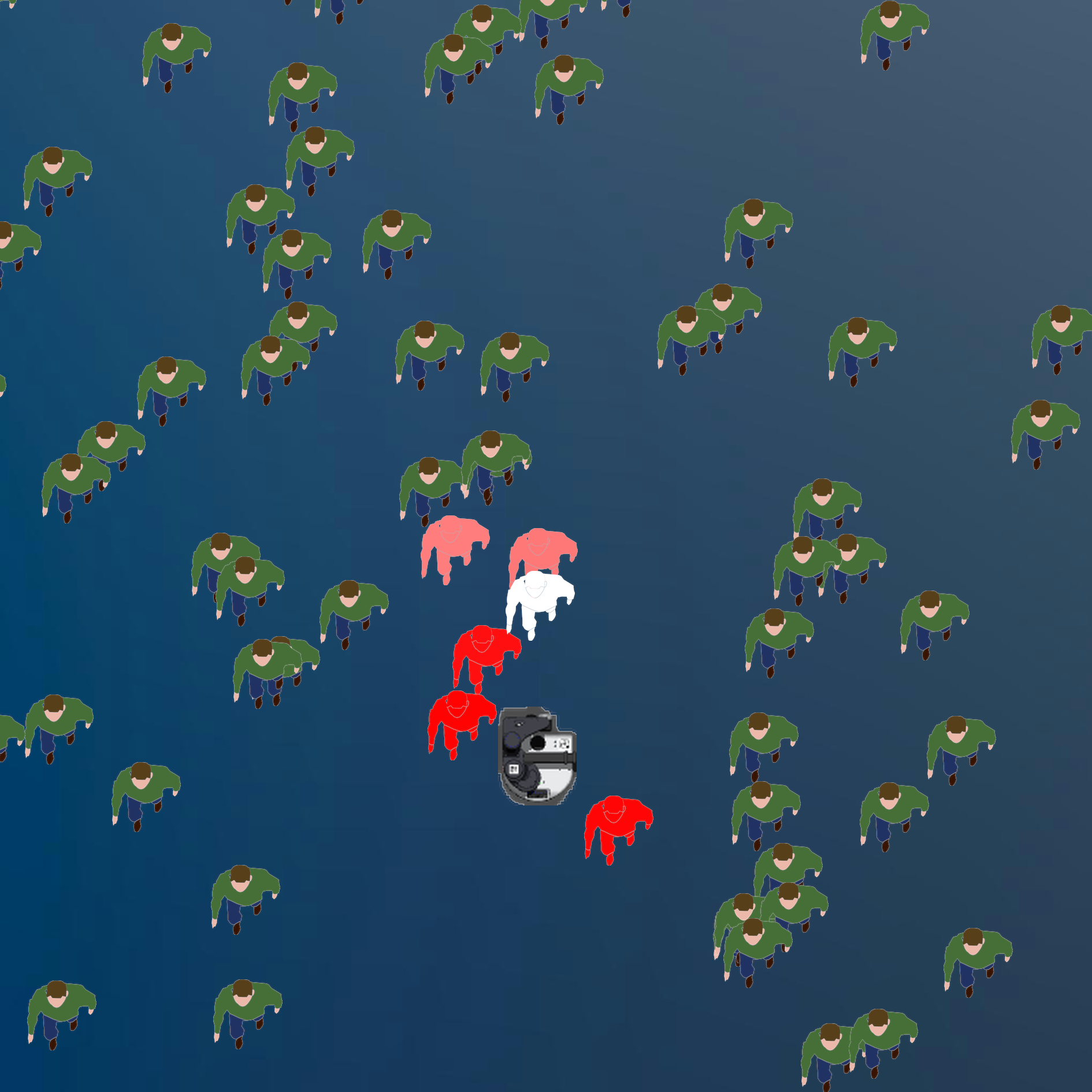}\\
    (a) MARS & (b) Navigation & (c) CrowdNav
  \end{tabular}
  \caption{The problem scenarios used to evaluate \solverAbbr.}
  \label{fig:example_problems}
\end{figure}

We tested \solverAbbr on three planning under uncertainty benchmark problems, detailed below. 

\subsection{Experimental Scenarios}

\noindent\textbf{Multi-Agent Rocksample (MARS)}~\cite{Pai21HyPdespot}: MARS$(n, m)$, shown in \Cref{fig:example_problems}(b) is an extension of the popular Rocksample benchmark problem, in which two agents (blue squares) operate in a $n\times n$ map populated by $m$ randomly placed rocks that are either \texttt{GOOD} (green rocks) or \texttt{BAD} (red rocks), resulting in $\left |\stSpace \right | = n^4 \times 2^m$. 
The agents do not know the rock states initially, but they are equipped with a noisy sensor to detect the state of a rock ($\left |\obsSpace \right | = 9$, including a \texttt{NULL} observation, when the sensor is not used). If an agent is on a rock, it can \texttt{SAMPLE} it (resuling in an action space of size $\left |\actSpace \right | = (5 + m)^2$), which yields a reward of $10$ for good rocks and $-10$ for bad rocks. Good rocks turn bad after sampling. The agents must work cooperatively to sample as many good rocks as possible, before leaving the map on the right-hand side (rewarded by $10$). The problem terminates when both agents leave the map or a maximum of $90$ planning steps has been reached. The discount factor in this problem is $\gamma=0.983$. More details of the problem can be found in~\cite{Pai21HyPdespot}.

\noindent\textbf{Navigation in a partially known map (Navigation)}~\cite{Pai21HyPdespot}: In this problem, shown in \Cref{fig:example_problems}(a), a robot (red square) starts from a random position at the top border of a map with $13 \times 13$ cells, consisting of randomly placed obstacles (black squares), and must reach a goal area (green square) at the bottom of the map by passing one of the two gates in the middle wall (blue squares), while avoiding collisions with the obstacles. The obstacles and which of the two gates is open are only partially known. The robot has access to a noisy sensor, providing information on which of the eight neighboring grid cells around the robot are occupied by an obstacle (resulting in $\left |\obsSpace \right | = 8$). In each step, the robot can move to one of its eight neighboring grid cells (resulting in $\left |\actSpace \right | = 8$) or remain in its current cell. Reaching the goal yields a reward of $20$, colliding with an obstacle and standing still yield a penalty of $-1$ and $-0.2$, respectively. Additionally, the robot receives a small penalty of $-0.1$ for every step. The problem terminates when the robot reaches the goal cell or after a maximum of $60$ planning steps. The discount factor is $\gamma=0.983$. This problem has a very large state space of size $\left |\stSpace \right | = 169 \times 2^{124}$, which includes the robot position and the occupancy of unknown cells. More details of the problem can be found in~\cite{Pai21HyPdespot}.

\noindent\textbf{Crowd Navigation (CrowdNav)}: 
We propose CrowdNav (\Cref{fig:example_problems}(c)), a problem in which a Stretch 3 mobile robot navigates in a conference hall of size $50\times40$m densely populated by $300$ randomly placed people. While the robot can fully observe the location of the people, their behavior is determined by an initially unknown character trait: each person is either curious with probability $p_{\text{curious}}$ or shy with probability $1-p_{\text{curious}}$. The state space consists of the robot's 2D position and the 2D positions and character traits of the $n$-nearest people to the robot (resulting in the state space $\stSpace = \mathbb{R}^2 \times \mathbb{R}^{2n} \times \{\texttt{CURIOUS}, \texttt{SHY}\}^n$). The motion of the people is stochastic. At each time step, the movement of all people is perturbed by zero-mean Gaussian noise with a standard deviation of $\sigma_p=0.05$. Additionally, with a probability of $0.9$, people within a radius $r_{\text{nearby}}$ of the robot also react based on their trait: curious ones move toward the robot with velocity $v_{\text{curious}}$, while shy ones move away with velocity $v_{\text{shy}}$. To navigate, the robot can choose from four directional actions—{\texttt{NORTH}, \texttt{EAST}, \texttt{SOUTH}, \texttt{WEST}}—each moving it one meter. It also has a \texttt{YELL} action (resulting in $\left |\actSpace \right | = 5$), which causes all nearby people to rapidly back away with velocity $v_{\text{back}}$. The robot's observation at each step encodes whether each of the $n$-nearest people decreased or increased their distance from its position (resulting in $\left |\obsSpace \right | = 2^n$). Since the crowd behavior is stochastic, this observation provides only an imperfect signal of their hidden traits. The robot's objective is to travel from the southern to the northern border of the hall, receiving a reward of $1,000$ for success. Bumping into a person incurs a penalty of $-200.$ Since using the \texttt{YELL} action might disturb nearby people, it incurs a penalty of $-25$. Additionally, each step incurs a small step penalty of $-1$. The problem terminates once the robot leaves the hall, or after a maximum of $200$ planning steps. In our experiments, we set $r_{\text{nearby}}=4m$, $v_{\text{curious}}=0.3m/s$, $v_{\text{shy}}=0.8m/s$, $v_{\text{back}}=2m/s$, and $n=6$. The discount factor is $\gamma=0.97$. 

\subsection{Experimental Setup}
The purpose of our experiments is two-fold: The first is to compare \solverAbbr with the state-of-the-art parallel online POMDP solver HyP-DESPOT~\cite{Pai21HyPdespot}, and the sequential solvers DESPOT~\cite{Ye2017despot} and POMCP~\cite{silver2010monte} in the Navigation and MARS problem scenarios. To do this, we implemented \solverAbbr and the problem scenarios in Python. We use PyTorch~\cite{paszke2019pytorch} as the backbone of \solverAbbr's tensor data structures and vectorized computations due to its maturity, simplicity, and rich API, though other libraries such as JAX~\cite{jax2018github} or Taichi~\cite{hu2019taichi} are possible, too. For \solverAbbr and POMCP, we first ran a set of systematic trials to determine the best parameters. For \solverAbbr, this includes the temperature parameter $\eta$ in \cref{eq:analytical_belief_value} and the number \numParallel of episodes that are sampled in parallel during our vectorized forward search (\Cref{ssec:forward_search}), which we set to $\eta = 2.0$ for both Navigation and MARS, and $\numParallel = 50,000$ and $\numParallel = 60,000$ for Navigation and MARS, respectively. For POMCP, this includes the UCB exploration constant, which we set to $20.0$ for Navigation and $5.0$ for MARS. For HyP-DESPOT and DESPOT, we use the implementation\footnote{\url{https://github.com/AdaCompNUS/hyp-despot}} and the parameters for both problem scenarios provided by the authors. The results of these experiments are presented in \Cref{ssec:comparison_w_despot}.


The second purpose is to demonstrate the efficacy of \solverAbbr in a challenging robotics planning under uncertainty problem. We tested \solverAbbr on the CrowdNav problem, where we investigated \solverAbbr's robustness to different crowd behaviors. Specifically, we varied the curiosity probability, $p_{\text{curious}}$, across five scenarios, where we used $p_{\text{curious}} \in \{0, 0.25, 0.5, 0.75, 1\}$. We then analyzed the robot trajectories computed by \solverAbbr in terms of length and safety. The results are presented in \Cref{ssec:crowd_nav_results}.

All experiments were carried out on the same machine\footnote{A laptop with one Intel Core i7-13850HX CPU with $32$GB of RAM and a Nvidia RTX 3500 ADA GPU with $12$GB of VRAM.}. \solverAbbr uses only the GPU. HyP-DESPOT uses both the CPU and GPU, while DESPOT and POMCP use only the CPU.

\subsection{Comparison with HyP-DESPOT, DESPOT and POMCP}\label{ssec:comparison_w_despot}

\begin{figure*}[htbp]
  \centering
  \begin{tabular}{@{}ccccc@{}}
    \includegraphics[width=0.16\textwidth]{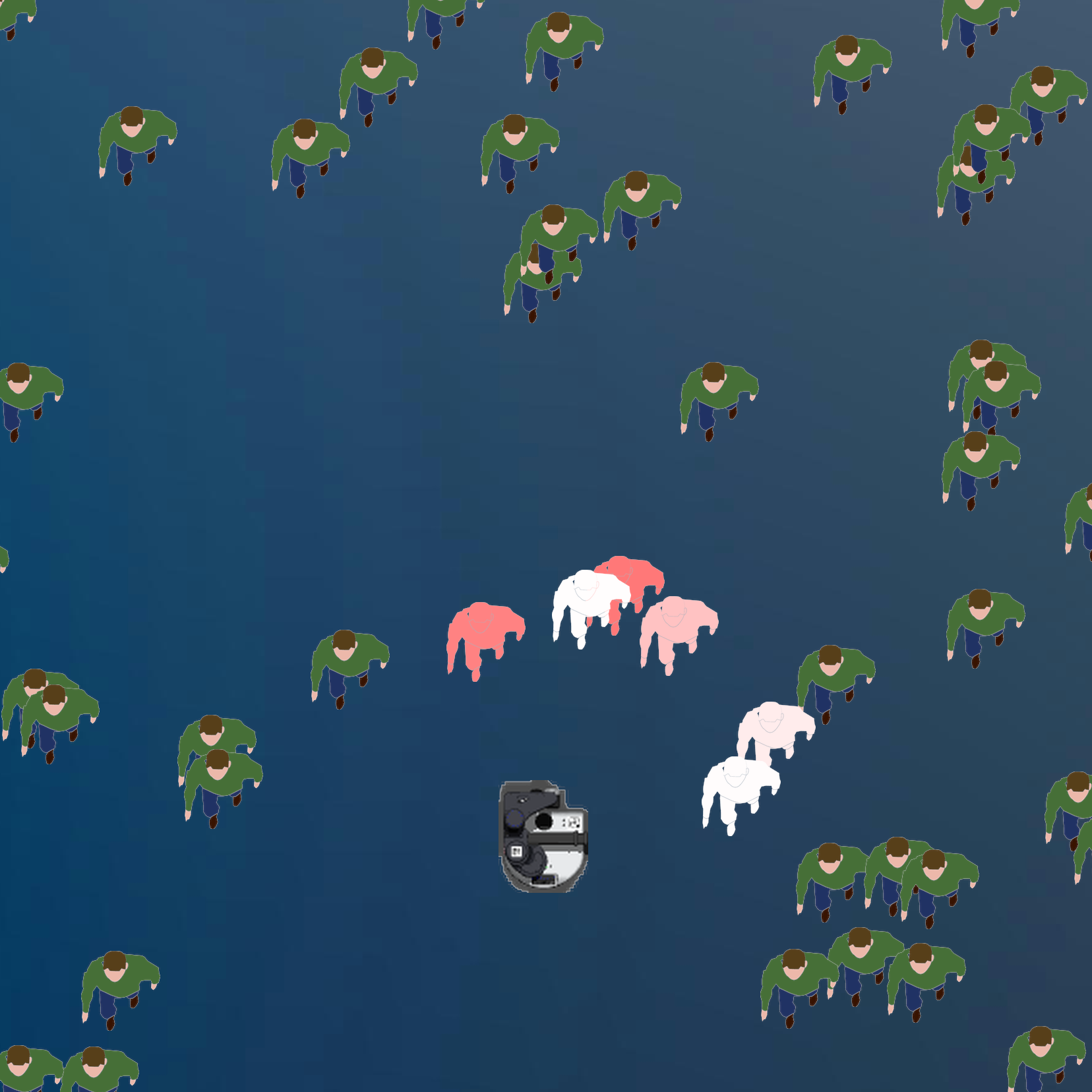} &
    \includegraphics[width=0.16\textwidth]{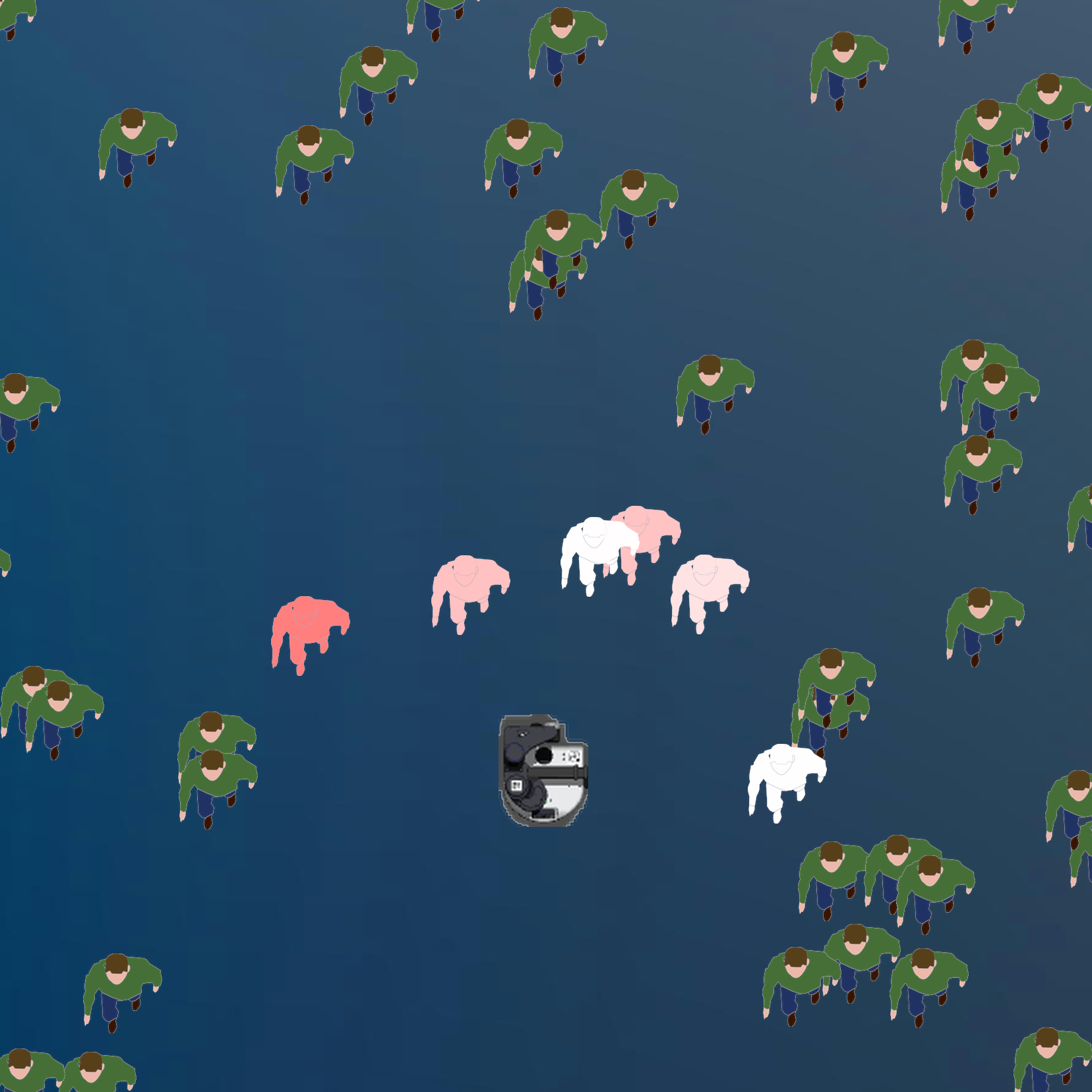} &
    \includegraphics[width=0.16\textwidth]{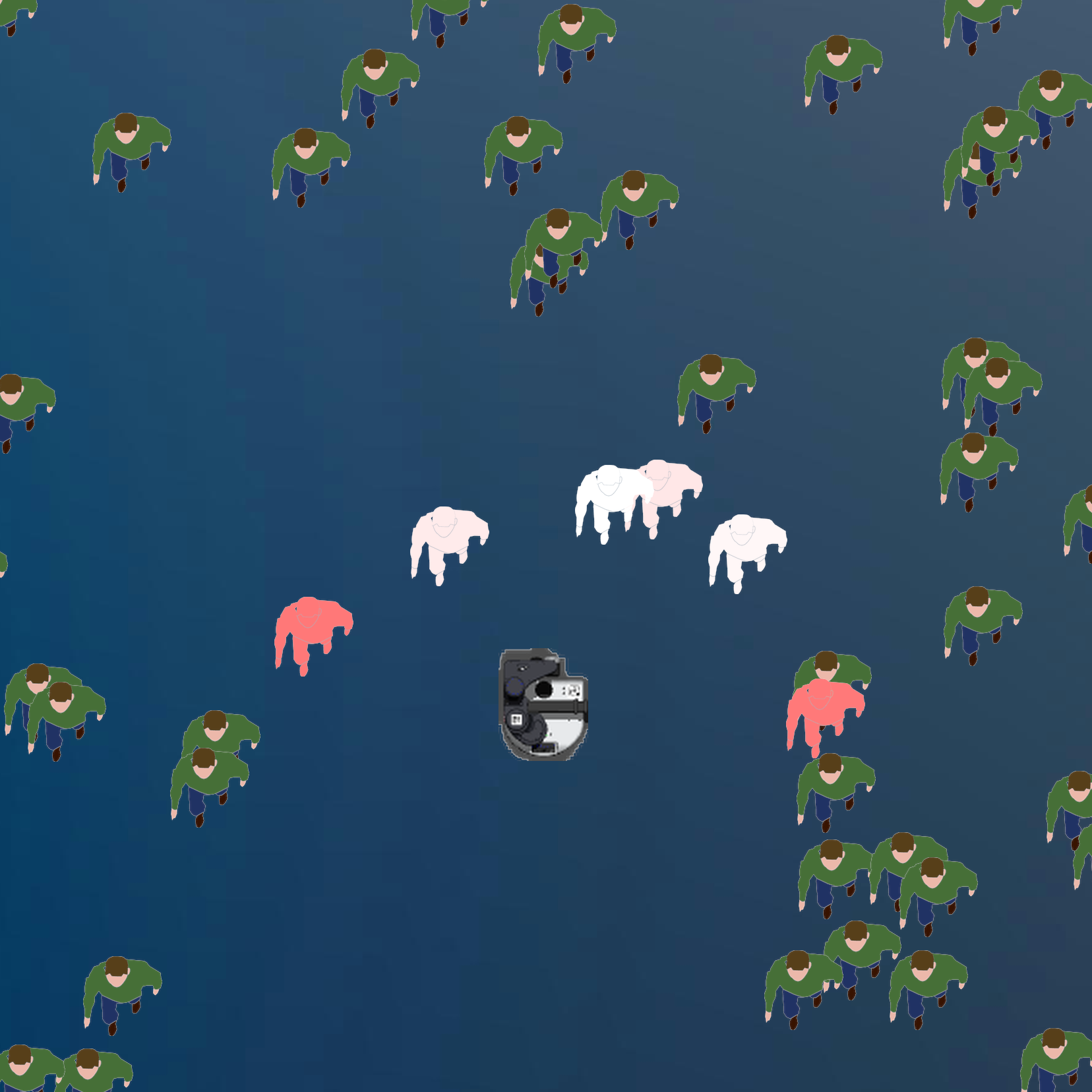} &
    \includegraphics[width=0.16\textwidth]{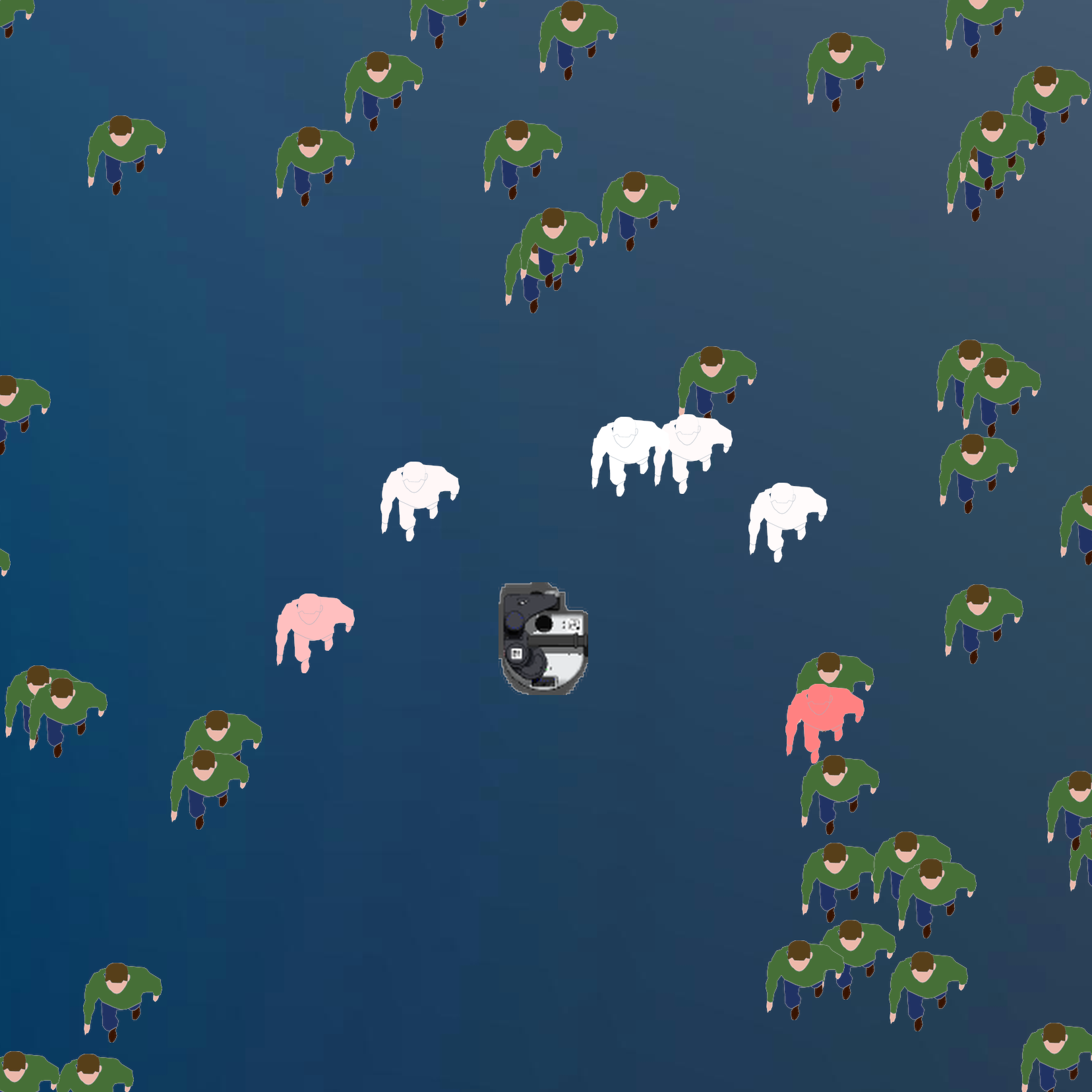} &
    \includegraphics[width=0.16\textwidth]{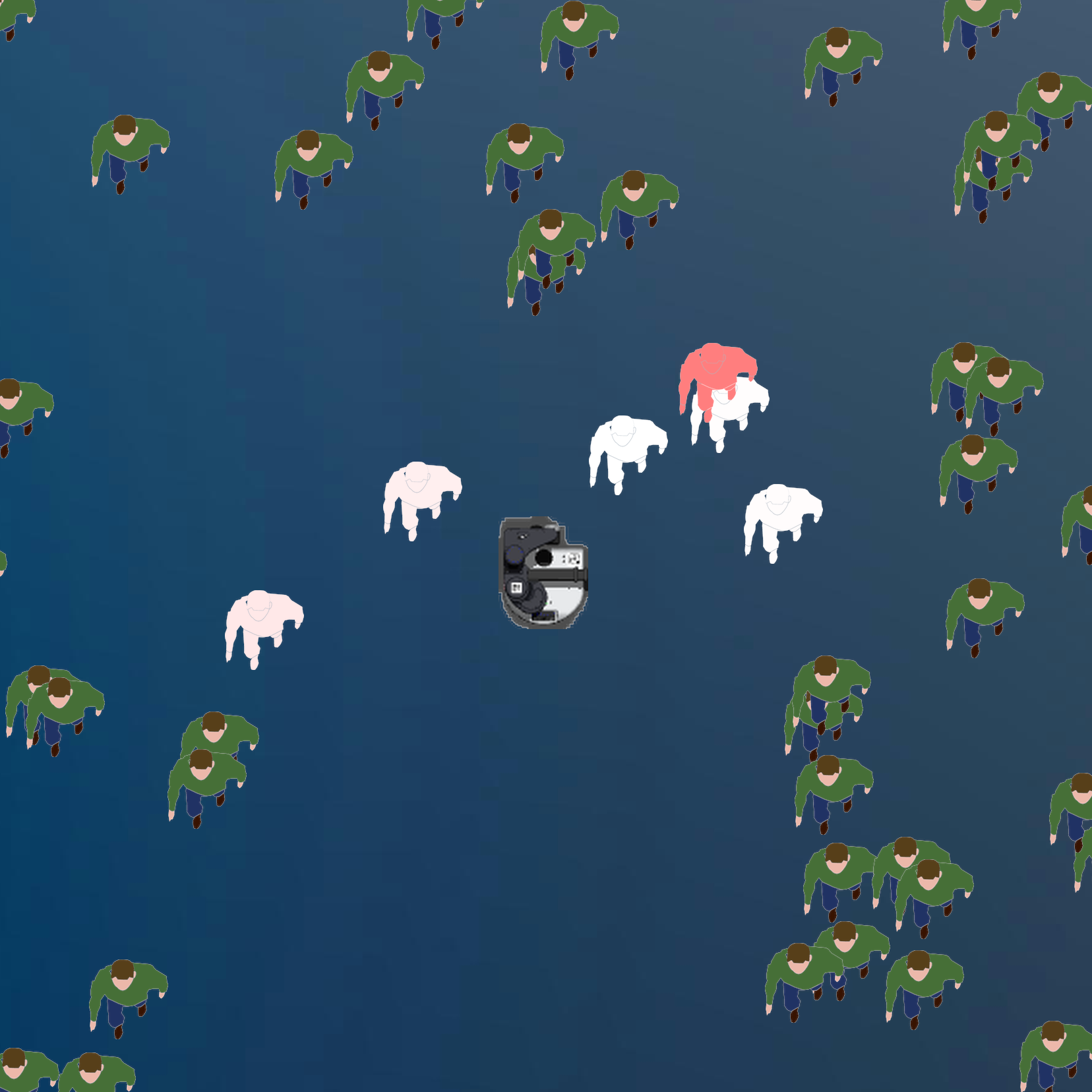} \\
    \addlinespace[0.4ex]\midrule\addlinespace[1.1ex]
    \includegraphics[width=0.16\textwidth]{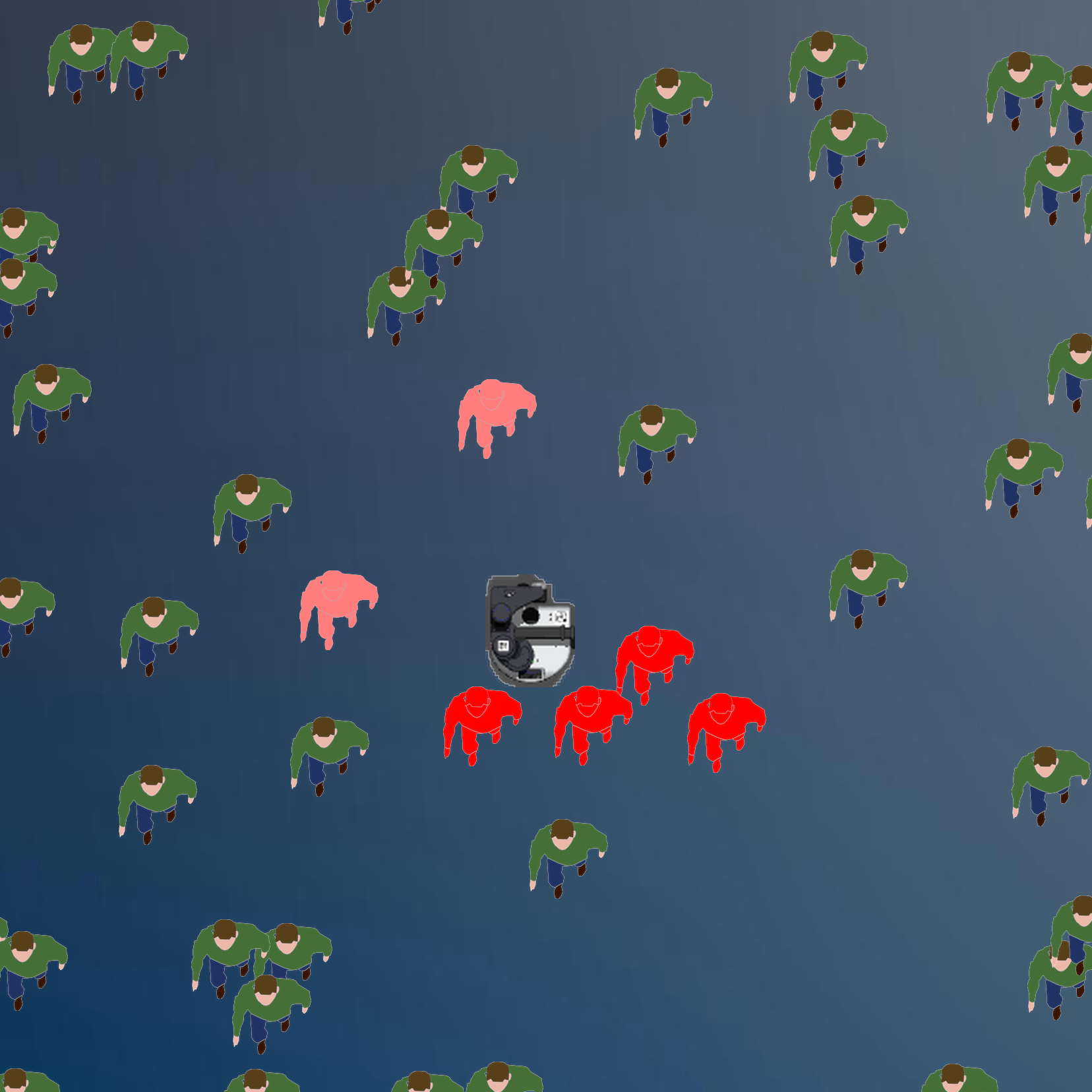} &
    \includegraphics[width=0.16\textwidth]{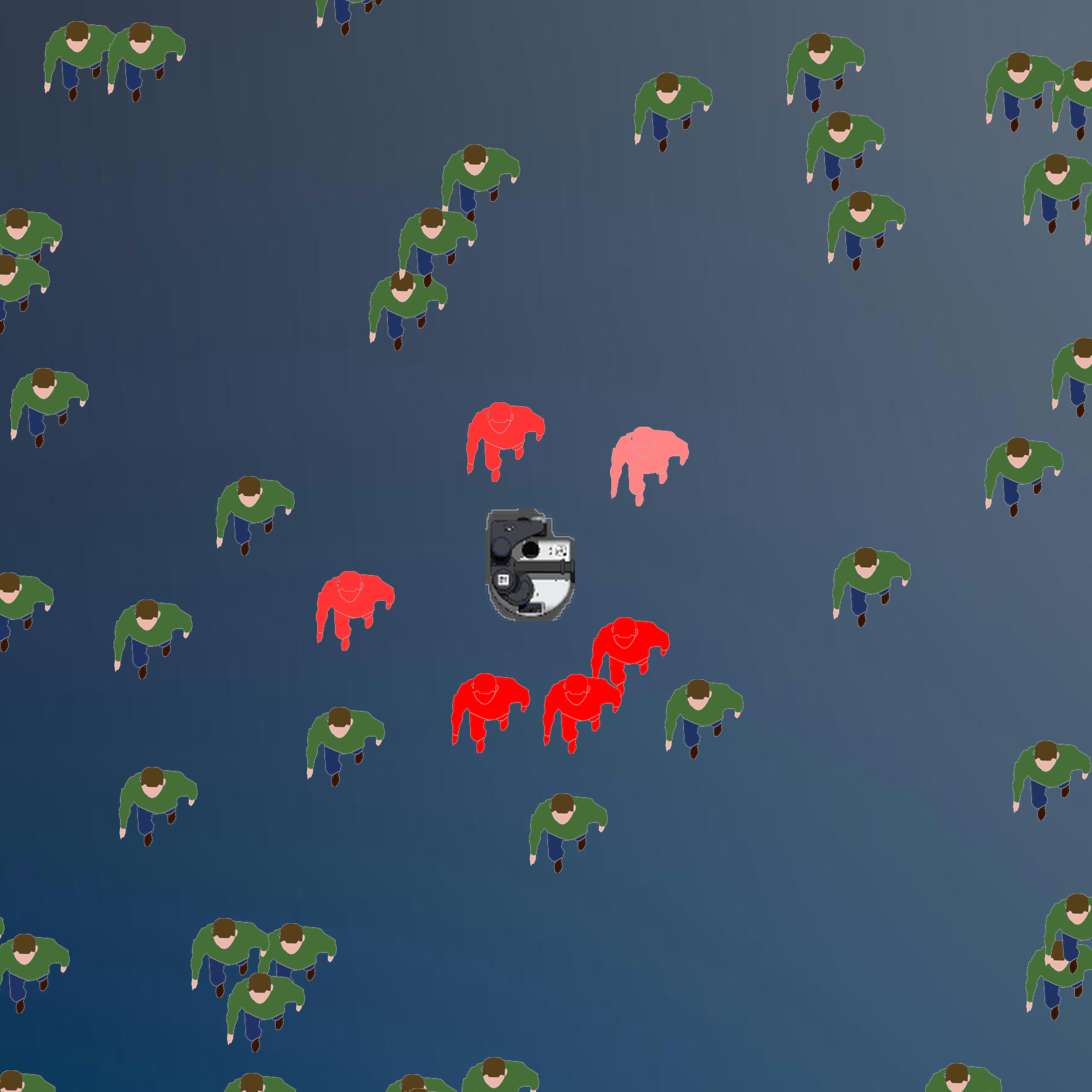} &
    \includegraphics[width=0.16\textwidth]{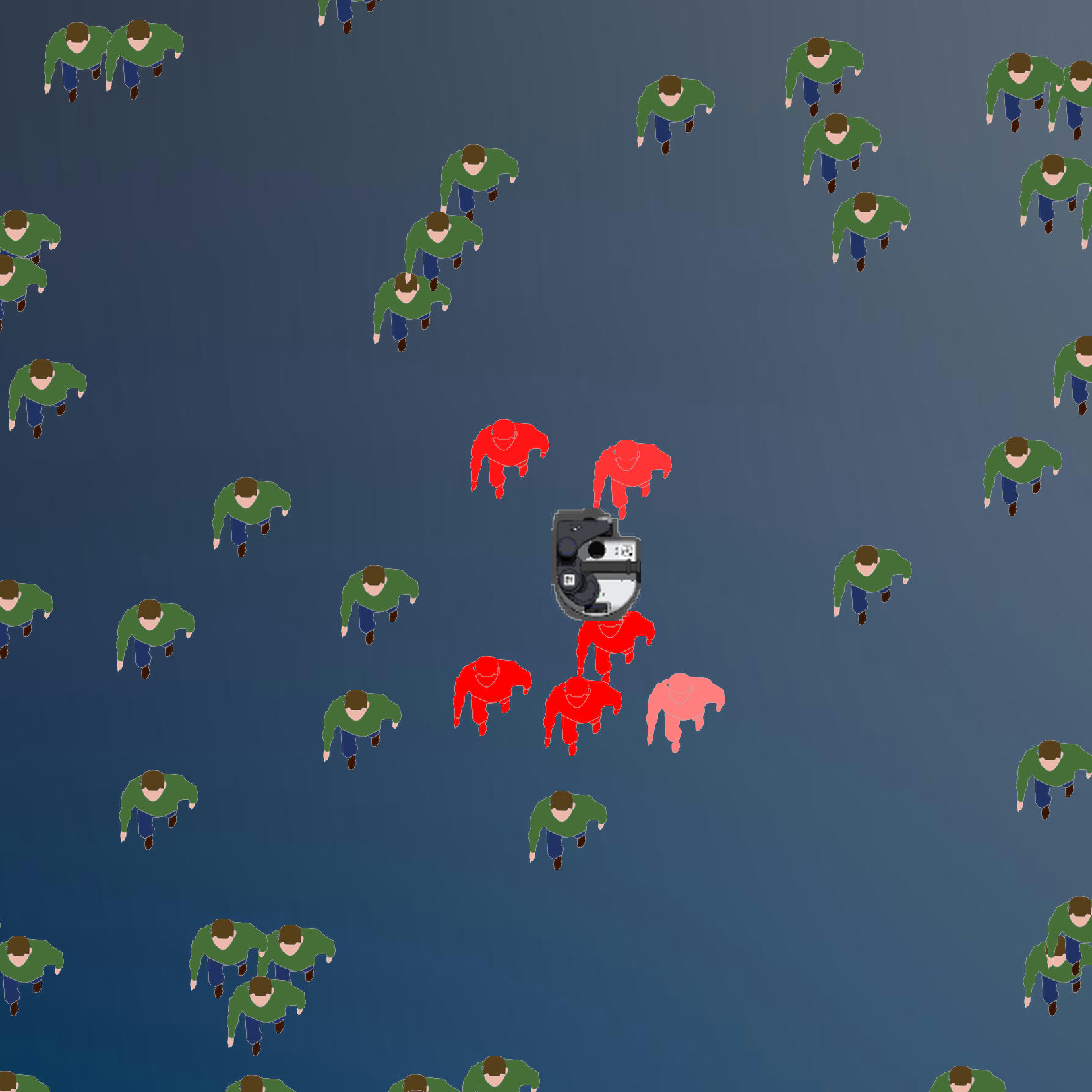} &
    \includegraphics[width=0.16\textwidth]{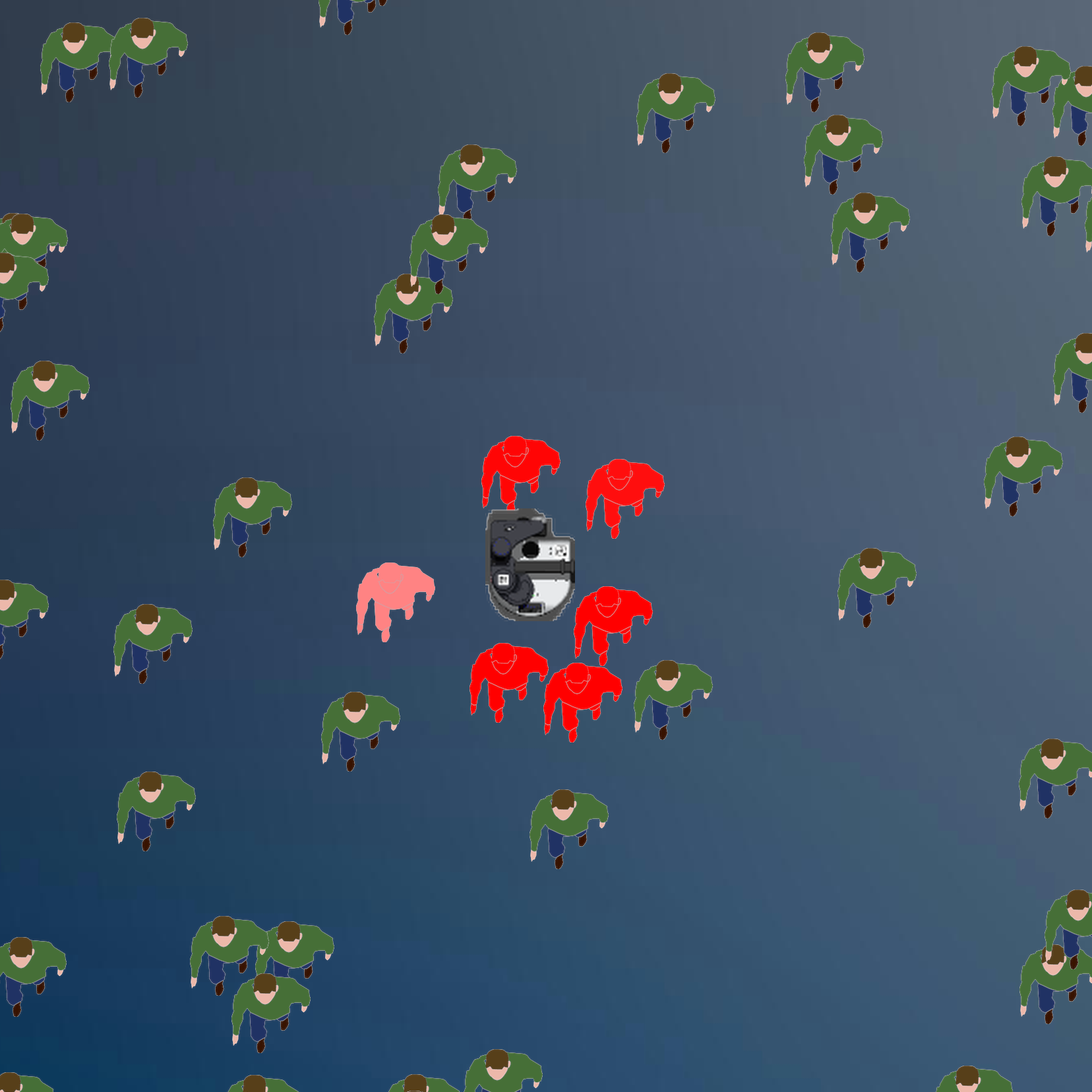} &
    \includegraphics[width=0.16\textwidth]{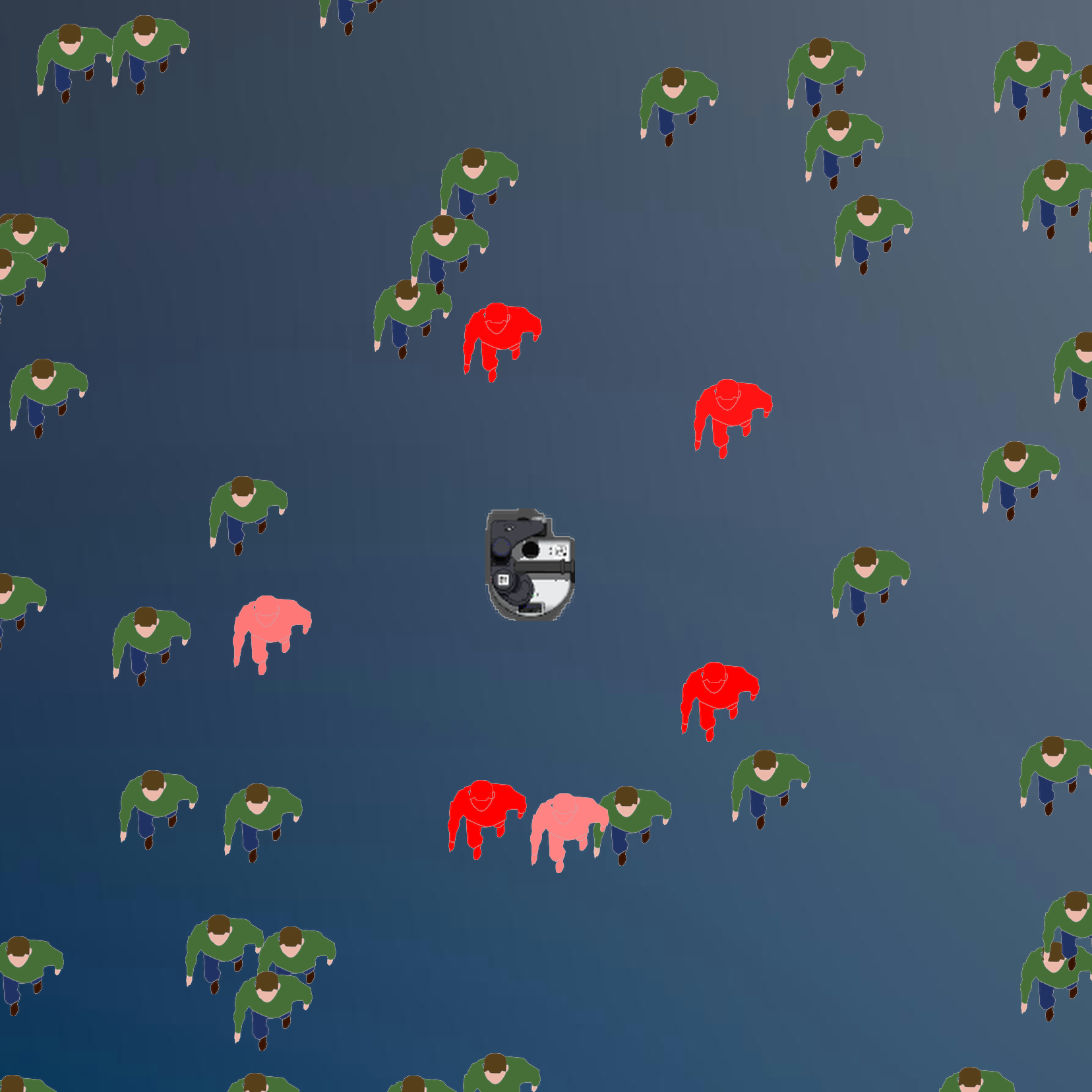} \\
    $t$ & $t+1$ & $t+2$ & $t+3$ & $t+4$
  \end{tabular}
  \caption{Two partial trajectories of the Stretch 3 mobile robot in the CrowdNav scenario with $p_{\text{curious}} = 0.0$ (top) and $p_{\textit{curious}} = 1.0$ (bottom) at different time steps. Nearby people are colored according to their inferred character trait. Darker red tones indicate a higher probability of a person being curious.
  }
  \label{fig:crowd_behaviors}
\vspace{-10pt}
\end{figure*}

\begin{table}[t]
\centering
\setlength{\tabcolsep}{3.5pt}   
\scriptsize
\caption{Average total discounted rewards and $95\%$ confidence intervals of all tested solvers on the Navigation \& MARS problems. Results are averaged over $200$ trials per solver, problem, and planning time per step.}
\label{t:comparison}
\begin{tabular}{@{}l|cccc@{}}
\hline\hline
 & \multicolumn{4}{c}{Planning time per step (s)} \\
\cline{2-5}
\textbf{Navigation}
& \rule{0pt}{2.0ex}$0.01$ & $0.05$ & $0.1$ & $1.0$ \\
\hline
\solverAbbr\ (Ours) 
& $\mathbf{8.8\pm1.0}$ 
& $\mathbf{10.7\pm0.8}$ 
& $\mathbf{11.7\pm0.7}$ 
& $\mathbf{11.9\pm0.6}$ \\
HyP-DESPOT 
& $3.1\pm1.4$ & $5.2\pm1.5$ & $5.7\pm1.6$ & $9.3\pm1.3$ \\
DESPOT & $0.3\pm1.9$ & $0.7\pm1.8$ & $2.2\pm1.8$ & $8.7\pm1.3$ \\
POMCP
& \rule{0pt}{2.0ex}$-2.1\pm0.9$ & $-2.0\pm0.8$ & $-1.9\pm1.1$ & $-0.5\pm1.0$ \\
\hline\hline
\textbf{MARS$(20,20)$} 
& \rule{0pt}{2.0ex}$0.01$ & $0.05$ & $0.1$ & $1.0$ \\
\hline
\solverAbbr\ (Ours) 
& $\mathbf{31.1\pm2.6}$ 
& $\mathbf{50.0\pm1.9}$ 
& $\mathbf{53.3\pm2.1}$ 
& $\mathbf{58.8\pm2.1}$ \\
HyP-DESPOT 
& $14.3\pm1.5$ & $19.2\pm2.0$ & $22.3\pm2.4$ & $47.9\pm1.6$ \\
DESPOT & $10.4\pm1.5$ & $17.8\pm2.1$ & $19.1\pm2.3$ & $20.9\pm2.0$ \\
POMCP
& $-588.3\pm73.1$ & $-9.7\pm3.5$ & $-5.9\pm3.0$ & $6.7\pm1.0$ \\
\hline
\hline
\multicolumn{5}{@{}l@{}}{
\parbox[t]{\columnwidth}{
\textbf{MARS$(50,50)$} (3025 actions, 1.0s planning time/step):\\
\solverAbbr\ achieved an average total discounted reward of $\mathbf{45.1\pm2.0}$ (200 trials).\\
HyP-DESPOT, DESPOT and POMCP crashed on MARS$(50, 50)$.
}
} \\
\hline \hline
\end{tabular}
\end{table}

\begin{table}[t]
\centering
\setlength{\tabcolsep}{4.25pt}
\scriptsize
\caption{Avg. number of steps to reach the goal and success rate in the Navigation problem, and the avg. percentage of good/bad rocks sampled in the MARS problems (relative to the total number of good/bad rocks per environment), averaged over $200$ trials.}
\label{t:comparison_metrics}
\begin{tabular}{l|cc|cc|cc}
\hline\hline
 & \multicolumn{2}{c|}{\textbf{Navigation}} & \multicolumn{2}{c|}{\textbf{MARS$(20,20)$}} & \multicolumn{2}{c}{\textbf{MARS$(50,50)$}}  \\
\cline{2-7}
 & Steps $\downarrow$ & Success (\%) $\uparrow$ 
 & Good $\uparrow$ & Bad $\downarrow$ & Good $\uparrow$ & Bad $\downarrow$ \\
\hline
\solverAbbr\ (Ours) 
& $\mathbf{19.8}$ & $\mathbf{94}$ 
& $\mathbf{90.0}$ & $2.0$ & $84.1$ & $5.1$ \\

HyP-DESPOT 
& $26.8$ & $77$ 
& $60.5$ & $\mathbf{1.2}$ & --- & --- \\

DESPOT 
& $27.9$ & $75$ 
& $33.4$ & $3.3$ & --- & --- \\

POMCP
& $53.3$ & $21$ 
& $11.2$ & $6.6$ & --- & --- \\

\hline\hline
\end{tabular}
\vspace{-6pt}
\end{table}


We first tested \solverAbbr, HyP-DESPOT and POMCP on MARS$(20, 20)$, a variant with a $20 \times 20$ map randomly populated by $20$ rocks. This variant has an action space of $625$ actions. We ran $200$ trials per solver, with planning times per step from $0.01$ to $1.0$s/step. \Cref{t:comparison} shows the average total discounted rewards achieved by the solvers. \solverAbbr clearly outperforms both HyP-DESPOT and POMCP by a significant margin accross all planning times/step. \solverAbbr computes strategies for which both agents tend to sample more good rocks before leaving the map (initially, the environment consists of $10$ good rocks on average), as shown in \Cref{t:comparison_metrics}.

The results further indicate that \solverAbbr is at least $20\times$ more efficient than HyP-DESPOT in this problem: For a planning time of $0.01$s/step, \solverAbbr achieves an average total discounted reward of $\mathord{\sim}64\%$ of what HyP-DESPOT achieves with $1$s/step. As we increase the planning time per step to $0.1$s, \solverAbbr achieves a better result than HyP-DESPOT with $1$s/step. In fact, the policies generated by \solverAbbr with $0.05$s planning time per step are better than what HyP-DESPOT can generate with $1$s/step.

We additionally evaluated DESPOT and POMCP on MARS$(20, 20)$ using a planning time of $10$s/step. DESPOT achieved an average total discounted reward of $27.9\pm4.8$, while POMCP achieved $10.3\pm4.4$ (averaged over $20$ trials). Both results are worse than what \solverAbbr achieves with only $0.01$s/step (see \Cref{t:comparison}). Thus, even with $1000\times$ larger planning budgets, the sequential solvers remain uncompetitive compared to \solverAbbr.

To test the scalability of \solverAbbr further, we ran $200$ trials on the MARS$(50, 50)$ problem, a variant of MARS with $3025$ actions. \solverAbbr handles this problem well, as indicated by the results in \Cref{t:comparison,t:comparison_metrics}. In contrast to many existing online solvers (including HyP-DESPOT), \solverAbbr does not require an exhaustive enumeration of all actions, which makes it much more suitable for solving POMDP problems with large action spaces. Unfortunately, we were unable to test HyP-DESPOT, DESPOT and POMCP on MARS$(50, 50)$, since the implementation provided by the authors crashed for variants larger than MARS$(36, 36)$.

For the Navigation problem, we tested all solvers using $200$ trials with varying planning times per step, from $0.01$ to $1.0$s/step. The results are shown in \Cref{t:comparison}. As in MARS, \solverAbbr significantly outperforms the comparators. This is also reflected in the average number of steps required to reach the goal and the success rate (percentage of runs for which the robot reaches the goal) in \Cref{t:comparison_metrics}.
\subsection{Navigation in a crowd}\label{ssec:crowd_nav_results}
\begin{table}[htb]
\setlength{\tabcolsep}{2.75pt}
\scriptsize
\caption{Average number of steps to reach the goal, number of times the robot bumped into people, and number of times the robot used the \texttt{YELL} action, together with $95\%$ confidence intervals, in the CrowdNav scenario for different values of $p_{\text{curious}}$. Results are averaged over $50$ trials for each $p_{\text{curious}}$.}\label{t:crowd_nav_res}
\vspace{-5pt}
\begin{tabular}{@{}l|ccccc@{}}
$p_{\text{curious}}$ & $0.0$ & $0.25$ & $0.5$ & $0.75$ & $1.0$ \\ \hline \hline
Steps & $77.2\pm2.7$ & $75.3\pm3.1$ & $90.7\pm5.6$ & $108.8\pm6.5$ & $123.6\pm7.8$ \\
Num. bumps & $0.0\pm0.0$ & $0.2\pm0.1$ & $0.4\pm0.1$ & $0.5\pm0.2$ & $0.8\pm0.2$ \\
Num. yells & $2.6\pm0.9$ & $2.4\pm0.5$ & $5.0\pm0.8$ & $8.1\pm1.0$ & $12.3\pm1.4$ \\
\end{tabular}
\end{table}

We tested \solverAbbr on the CrowdNav problem using $50$ trials for each curiosity probability $p_{\text{curious}} \in \{0, 0.25, 0.5, 0.75, 1\}$ with a maximum planning time of $1$s/step. 

The results are summarized in \Cref{t:crowd_nav_res}. In all trials, the robot reached its goal at the northern border of the hall. With increasing $p_{\text{curious}}$, crowds consists of more curious people, causing the robot to take detours in order to avoid bumping into people. This leads to longer paths while the number of collisions increases only marginally.

More interestingly, different crowd behaviors (in terms of the curiosity probability $p_{\text{curious}}$) lead to vastly different strategies based on the inferred character traits of nearby people. For small $p_{\text{curious}}$, the crowd consists mostly of shy people who avoid the robot. Over time, the robot infers this character trait and adapts its strategy by taking more direct actions towards the goal. An example is shown in \Cref{fig:crowd_behaviors} (top row), where the surrounding people are inferred to be shy with high probability. As a result, the robot dashes towards the goal, even if the direct path is currently blocked. For large $p_{\text{curious}}$, the robot is often surrounded by curious people. In such situations, the robot uses the \texttt{YELL} action, causing nearby people to temporarily retreat from the robot. An example of this behavior is shown in \Cref{fig:crowd_behaviors} (bottom row). The nearby people are inferred to be curious with high probability and surround the robot at time step $t+3$. The robot then uses the \texttt{YELL} action, helping it avoid collisions.



\section{CONCLUSION}
We propose a novel parallel online POMDP solver, called \solverAbbr, the first fully vectorized online solver. \solverAbbr builds on a recent POMDP formulation that analytically solves value functions and only leaves the estimation of expectations for numerical computations, thereby removing any requirements for explicit synchronization between parallel computations. \solverAbbr represents all data structures related to the belief tree as tensors and formulates planning as a sequence of fully vectorized operations over this representation, with no dependencies between parallel computations. This allows \solverAbbr to fully leverage the massive data parallel throughput of modern GPUs. Experimental results indicate that \solverAbbr is at least $20\times$ more efficient than a current state-of-the-art parallel online POMDP solver. Furthermore, \solverAbbr outperforms state-of-the-art sequential online solvers, while using a $1000\times$ smaller planning budget.




\bibliographystyle{IEEEtran}
\bibliography{IEEEabrv,references}

\end{document}